\documentclass[11pt]{article}
\usepackage{eacl2017}
\usepackage{times}
\usepackage{url}
\usepackage{tabularx}
\usepackage{latexsym}

\eaclfinalcopy 
\def\eaclpaperid{436} 

\usepackage{algorithm}
\usepackage[noend]{algpseudocode}
\algrenewcommand{\algorithmicindent}{1em}
\usepackage{mathtools}
\usepackage{paralist}
\usepackage{txfonts}
\usepackage{tikz}
\usetikzlibrary{automata,arrows,shapes}
\tikzset{every edge/.style={draw,->,>=stealth',shorten >=1pt,semithick}}
\tikzset{label/.style={sloped,above,pos=0.5,font={\tiny}}}
\tikzset{initial text={},double distance=2pt}
\usetikzlibrary{matrix}
\newcommand{\bos}{\texttt{<s>}}
\newcommand{\eos}{\texttt{</s>}}
\newcommand{\forward}{\ensuremath{\alpha}}
\newcommand{\nnz}{\ensuremath{z}}

\usepackage{graphicx}
\usepackage{color}

\DeclareMathOperator{\logonep}{log1p}

\newcommand{\pluseq}{\mathrel{\mathord+\mathord=}}

\title{Decoding with Finite-State Transducers on GPUs}

\author{Arturo Argueta \and David Chiang \\ Department of Computer Science and Engineering \\ University of Notre Dame \\ {aargueta,dchiang}@nd.edu}

\date{}

\begin{document}
\maketitle
\begin{abstract}
Weighted finite  automata and transducers (including hidden Markov models and conditional random fields) are widely used in natural language processing (NLP) to perform tasks such as morphological analysis, part-of-speech tagging, chunking, named entity recognition, speech recognition, and others. 
Parallelizing finite state algorithms on graphics processing units (GPUs) would benefit many areas of NLP. Although researchers have implemented GPU versions of basic graph algorithms, limited previous work, to our knowledge, has been done on GPU algorithms for weighted finite automata. We introduce a GPU implementation of the Viterbi and forward-backward algorithm, achieving decoding speedups of up to 5.2x over our serial implementation running on different computer architectures and 6093x over OpenFST.
\end{abstract}

\section{Introduction}

Weighted finite automata \cite{mohri:2009}, including hidden Markov models and conditional random fields \cite{lafferty+al:2001}, are used to solve a wide range of natural language processing (NLP) problems, including phonology and morphology, part-of-speech tagging, chunking, named entity recognition, and others. 
Even models for speech recognition and phrase-based translation can be thought of as extensions of finite automata \cite{mohri+al:2002,kumar+al:2005}.

Although the use of graphics processing units (GPUs) is now \emph{de rigeur} in applications of neural networks and made easy through toolkits like Theano \cite{theano}, there has been little previous work, to our knowledge, on acceleration of weighted finite-state computations on GPUs \cite{narasiman2011,li2014,peng2016,chong2009f}. In this paper, we consider the operations that are most likely to have high speed requirements: decoding using the Viterbi algorithm, and training using the forward-backward algorithm. We present an implementation of the Viterbi and forward-backward algorithms for CUDA GPUs. We release it as open-source software, with the hope of expanding in the future to a toolkit including other operations like composition.

Most previous work on parallel processing of finite automata \cite{ladner+fischer:1980,hillis,Mytkowicz} uses dense representations of finite automata, which is only appropriate if the automata are not too sparse (that is, most states can transition to most other states). But the automata used for natural language tend to be extremely large and sparse.  In addition, the more recent work in this line 
assumes deterministic automata, but automata that model natural language ambiguity are generally nondeterministic. 

Previous work has been done on accelerating particular NLP tasks on GPUs: in machine translation, phrase-pair retrieval \cite{he2013} and language model querying \cite{bogoychevn}; parsing \cite{hall,canny2013}; and speech recognition \cite{kim2012}. Our aim here is for a more general-purpose collection of algorithms for finite automata.

Our work uses concepts from the work of \newcite{Merril}, who show that GPUs can be used to  accelerate breadth-first search in sparse graphs. 
Our approach is simple, but well-suited to the large, sparse automata that are often found in NLP applications. 
We show that it achieves a speedup of a factor of 5.2 on a GPU relative to a serial algorithm, and 6093 relative to OpenFST.

\section{Graphics Processing Units}
GPUs became known for their ability to render high quality images faster than conventional multi-core CPUs. Current off-the-shelf CPUs contain 8--16 cores while GPUs contain 1500--2500 simple CUDA cores built into the card. General Purpose GPUs (GPGPU) contain cores able to execute calculations that are not constrained to image processing. GPGPUs are now widely used across scientific domains to enhance the performance of diverse applications.

\subsection{Architecture}
CUDA cores (also known as scalar processors) are grouped into different Streaming Multiprocessors (SM) on the graphics card. The number of cores per SM varies depending on the GPU's micro-architecture, ranging from 8 cores per SM (Tesla) up to 192 (Kepler). The overall number of SM on the chip varies, and it can range from 15 (Kepler) up to 24 (Maxwell). Streaming Multiprocessors are composed of the following components:
\begin{itemize}
    \item \textbf{Special Function units} (SFU) These allow computations of functions such as sine, cosine, etc.
    \item \textbf{Shared Memory and L1 Cache} The size of the memory varies on the GPU model.
    \item \textbf{Warp Schedulers} assigns threads in an SM to be executed in a specific warp.
\end{itemize}
To execute a workload on the GPU, a kernel must be launched with a specified grid structure. The kernel must specify the number of threads to run on a block and the number of blocks in a grid before being executed on the device. The maximum number of threads per block and blocks per grid can vary depending on the GPU device. If the kernel is successfully launched, each block in the grid will get assigned to a SM. Each SM will execute 32 threads at a time (also called a warp) in its assigned block. If the number of threads in a block is not divisible by 32, the kernel will not launch on the device. Each SM contains a warp scheduler in charge of choosing the warps in a block to be executed in parallel. When the amount of blocks in a grid surpasses the amount of SM on the device, the SMs will execute a subset of blocks in parallel.\\

\begin{table}
\centering
\begin{tabular}{|c|c|}
 \hline
 \multicolumn{2}{|c|}{K40 Specs} \\
 \hline
 Global Memory & 11520 MB\\
 \hline
  L2 cache size & 1.57 MB\\
 \hline
 Shared memory per block & 0.049 MB\\
 \hline
 Multiprocessors & 15\\
 \hline
 Cores per MP & 192\\
 \hline  
 Registers per block & 65536\\
 \hline
\end{tabular}
\caption{Device properties of a K40c GPU}
\label{fig:gpu_specs}
\end{table}

The memory hierarchy on the device is laid out to maximize the data throughput. Table \ref{fig:gpu_specs} shows the amount of cores available for execution as well as the amount of memory available on a Kepler based GPU. Registers are the fastest type of memory on the device, and this memory is private to each thread running on a block. Shared memory is the second fastest, and is shared by all threads running in the same block. The next type of memory is the L2 cache, which is shared among all streaming multiprocessors. The slowest and largest type of memory is global memory. Directly reading and writing to global memory affects performance significantly. Efficient memory management (reading and writing to and from contiguous addresses in memory) is important to fully utilize the memory hierarchy and increase performance.

\subsection{Optimizations}
\label{sec:optimizations}
Different factors such as number of threads in a block or coalesced memory accesses affect the performance on the GPU. In this section, we will cover the methods and modifications we used to improve the performance of our parallel implementations.

The optimal number of threads per block depends on the device configuration. The number of multiprocessors and cores per multiprocessor must be considered before launching a CUDA kernel on the device. Table \ref{fig:gpu_specs} shows the number of streaming multiprocessors and the number of cores per multiprocessor on a K40 GPU. Multiple blocks in a kernel grid can get scheduled to be executed on a single streaming multiprocessor if the number of blocks in a grid exceeds the number of streaming multiprocessors. Each streaming multiprocessor will only execute one warp in a block in parallel during execution, and that is why choosing an appropriate number of blocks is important. For example, if two blocks get assigned to a multiprocessor and each block contains 192 threads, the multiprocessor must execute 12 warps total where 1 warp gets executes at a time in parallel.

In our implementations, we take the following approach.
The number of cores per multiprocessor is considered first to configure the block size. The block size  is set to contain the same number of threads as the number of cores per multiprocessor of the graphics card used. If the number of threads needed to perform a computation is not divisible by the amount of cores per multiprocessor, the number of threads is rounded up to the closest dividend. Once the block size and number of threads are selected, the number of blocks is chosen by dividing the total number of threads by the block size.

Coalesced memory accesses are essential to maximize the use of resources running on the GPU. When data is requested by a warp executing on a streaming multiprocessor, a block from global memory will be accessed and allocated in shared memory. It is crucial to coalesce memory accesses so the number of blocks of global memory requested and the global memory access times decrease. This can be achieved by making all threads in a warp access contiguous spaces in memory. A similar speedup can be achieved if each thread in a block allocates all the data required from global memory into a compact data structure allocated in shared memory (size of the shared memory varies across devices). Section \ref{serial_algo} describes the data structure used to coalesce memory reads. For each input symbol $w_t$ the source states of all possible transitions can be read in a coalesced form and stored in shared memory allowing faster execution times.

Using special function units on the device can inhibit the performance of a program running on the GPU. Performance is affected because the number of SFU is lower than the amount of regular cores (e.g. The GK104 Kepler architecture contains 1536 regular cores and 256 special function units total). Also, the cycle penalty for using SFU rather than CUDA cores is higher than the penalty for regular cores on the device. For this work, the amount of instructions that use a specific SFU are kept to a minimum to obtain a higher speedup. By combining the mentioned techniques in this section, an application can significantly increase its performance. 

\section{Weighted Finite Automata}
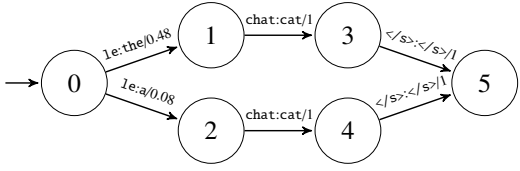
\begin{figure}
\begin{center}
\scalebox{0.9}{%
\begin{tikzpicture}[y=20pt]
\node[initial,state](0) {0};
\node[state](1) at (2,1) {1};
\node[state](2) at (2,-1) {2};
\node[state](3) at (4,1) {3};
\node[state](4) at (4,-1) {4};
\node[state](5) at (6,0) {5};
\draw (0) edge node[label] {\texttt{le}:\texttt{the}/0.48} (1);
\draw (0) edge node[label] {\texttt{le}:\texttt{a}/0.08} (2);
\draw (1) edge node[label] {\texttt{chat}:\texttt{cat}/1} (3);
\draw (2) edge node[label] {\texttt{chat}:\texttt{cat}/1} (4);
\draw (3) edge node[label] {\eos:\eos/1} (5);
\draw (4) edge node[label] {\eos:\eos/1} (5);
\end{tikzpicture}}
\end{center}
\caption{Example of a FST that translates the french string \texttt{le chat} to English.}
\label{fig:fst_example}
\end{figure}

In this section, we review weighted finite automata, using a matrix formulation. A \emph{weighted finite automaton} is a tuple $M = (Q, \Sigma, s, F, \delta)$, where
\begin{itemize}
\item $Q$ is a finite set of states.
\item $\Sigma$ is a finite input alphabet.
\item $s \in \mathbb{R}^Q$ is a one-hot vector: if $M$ can start in state $q$, then $s[q]=1$; otherwise, $s[q]=0$. 
\item $f \in \mathbb{R}^Q$ is a vector of final weights: if $M$ can accept in state $q$, then $f[q]>0$ is the weight incurred; otherwise, $f[q]=0$.
\item $\delta : \Sigma \rightarrow \mathbb{R}^{Q \times Q}$ is the transition function: if $M$ is in state $q$ and the next input symbol is $a$, then $\delta[a][q,q']$ is the weight of going to state~$q'$.
\end{itemize}
Note that we currently do not allow transitions on empty strings or epsilon transitions.
This definition can easily be extended to weighted finite transducers by augmenting the transitions with output symbols. 
See Figure~\ref{fig:fst_example} for an example FST.

Using this notation, the total weight of a string $w=w_1\cdots w_n$ can be written succinctly as:
\begin{equation}
\mathsf{weight}(w) = s^\top \, \left(\prod_{t=1}^n \delta[w_t] \right)\, f.
\label{eq:weight}
\end{equation}
Matrix multiplication is defined in terms of multiplication and addition of weights. It is common to redefine weights and their multiplication/addition to make the computation of (\ref{eq:weight}) yield various useful values. When this is done, multiplication is often written as~$\otimes$ and addition as~$\oplus$. If we define $p_1 \otimes p_2 = p_1p_2$ and $p_1 \oplus p_2 = p_1+p_2$, then equation (\ref{eq:weight}) gives the total weight of the string. 

Or, we can make Equation (\ref{eq:weight}) obtain the \emph{maximum} weight path as follows. The weight of a transition is $(p, k)$, where $p$ is the probability of the transition and $k$ is (a representation of) the transition itself. Then
\begin{align*}
(p_1, k_1) \otimes (p_2, k_2) &\equiv (p_1p_2, k_1 k_2) \\
(p_1, k_1) \oplus (p_2, k_2) &\equiv \begin{cases}
(p_1, k_1) & \text{if $p_1 > p_2$} \\
(p_2, k_2) & \text{otherwise.} \\
\end{cases}
\end{align*}
The Viterbi algorithm simply computes Equation (\ref{eq:weight}) under the above definition of weights.

\section{Serial Algorithm}
\label{serial_algo}

Applications of finite automata use a variety of algorithms, but the most common are the Viterbi, forward, and backward algorithms. Several of these automata algorithms are related to one another and used for learning and inference. Speeding up these algorithms will allow faster training and development of large scale machine learning systems.

The forward and backward algorithms are used to compute weights (Eq. \ref{eq:weight}), in left-to-right (Reading an input utterance from left to right) and right-to-left order, respectively. Their intermediate values are used to compute expected counts during training by expectation-maximization \cite{eisner:2002}. They can be computed by Algorithm \ref{alg:forward_backward_csr}.

Algorithm~\ref{alg:forwardcoo} is one way of computing Viterbi using Equation~(\ref{eq:weight}). It is a straightforward algorithm, but the data structures require a brief explanation.

Throughout this paper, we use zero-based indexing for arrays.
Let $m=|\Sigma|$, and number the input symbols in $\Sigma$ consecutively $0, \ldots, m-1$.
Then we can think of $\delta$ as a three-dimensional array. In general, this array is very sparse. We store it using a combination of compressed sparse row (CSR) format and coordinate (COO) format, as shown in Figure~\ref{fig:coo} where:

\begin{itemize}
    \item $\nnz$ is the number of transitions with nonzero weight
    \item $R$ is an array of length $(m+1)$ containing offsets into the arrays $S$,$T$,$O$, and $P$. if $a \in \Sigma$, the transitions on input $a$ can be found at positions
$R[a], \ldots R[a+1]-1$ (i.e. to access all transitions  $\delta[a]$ ). Note that $R[m] = \nnz$
    \item $S$ contains the source states for each transition $0 \leq k < \nnz \in  \delta[a]$
    \item $T$ contains target states for transitions $0 \leq k < \nnz \in  \delta[a]$
    \item $O$ contains the output symbols for transitions from state $S[k]$ to state $T[k]$
    \item $P$ contains the probabilities for transitions from state $S[k]$ to state $T[k]$
\end{itemize}

 The vector $f$ of final weights is stored as a sparse vector: for each $k$, $S_f[k]$ is a final state with weight~$P_f[k]$.

\begin{figure}
\begin{center}
\begin{tikzpicture}
\tikzset{every matrix/.style={matrix of nodes,
        anchor=west,
        line width=0.4pt,column sep=-0.4pt,
        nodes={
          rectangle,draw=black,minimum width=2em,text height=1.1em,text depth=0.5em,font=\footnotesize,inner sep=0},
        column 1/.style={nodes={draw=none}  }}}
\matrix[row 1/.style={nodes={draw=none}}] (r) at (0,1in) { 
    & le & chat & \eos \\
$R$ & 0 & 2 & 4 & 6 \\
};
\matrix [row sep=4pt] (m) at (0,0) { 
$S$ & 0 & 0 & 1 & 2 & 3 & 4 \\
$T$ & 1 & 2 & 3 & 4 & 5 & 5 \\
$O$ & the & a & cat & cat & \eos & \eos \\
$P$ & 0.48 & 0.08 & 1 & 1 & 1 & 1 \\
};

\tikzset{->}
\draw (r-2-2.south) -- (m-1-2.north);
\draw (r-2-3.south) -- (m-1-4.north);
\draw (r-2-4.south) -- (m-1-6.north);

\end{tikzpicture}
\\

\end{center}
\caption{CSR/COO representation of FST in Figure~\ref{fig:fst_example}. 
}
\label{fig:coo}
\end{figure}

\begin{algorithm}
\begin{algorithmic}[1]
\For{$q \in Q$}
  \State{$\forward[0][q] = 0$}
\EndFor
\State{$\forward[0][s] = 1$}
\For{$t = 1, \ldots, n$}
  \State{$a \leftarrow w_t$}
  \For{$k = R[a], \ldots, R[a+1]-1$} \label{line:kloop}
  \State{$p \leftarrow \forward[t-1][S[k]] \otimes P[k]$} \label{line:prob}
  \State{$\forward[t][T[k]] \leftarrow \forward[t][T[k]] \oplus p$} \label{line:max}
  \EndFor
\EndFor
\State{\textbf{return}  $\bigoplus_k \forward[n][S_f[k]] \otimes P_f[k] $}
\end{algorithmic}
\caption{Serial Viterbi algorithm (using CSR/COO representation).}
\label{alg:forwardcoo}
\end{algorithm}

\begin{algorithm}
\begin{algorithmic}[1]
\State{$\mathsf{forward}[0][s] \leftarrow 1$} \Comment{Begin forward pass}
\For{$t=0, \ldots, n-1$}
  \For{$q \in Q$}
    \For{$q' \in Q$ such that $\delta[w_{t+1}][q,  q']>0$}
      \State{$p = \mathsf{forward}[t][q]  \delta[w_{t+1}][q, q']$}
      \State{$\mathsf{forward}[t+1][q']\pluseq p$}\label{line:update1}
    \EndFor
  \EndFor
\EndFor

\For{$q \in Q$} \Comment{backward pass}
\State{$\mathsf{backward}[n][q] = f[q]$}
\EndFor
\For{$t=n-1, \ldots, 0$}
\For{$q \in Q$}
  \For{$q' \in Q$ such that $\delta[w_{t+1}][q, q']>0$}
      \State{$p = \delta[w_{t+1}][q, q']  \mathsf{backward}[t][q']$}
      \State{$\mathsf{backward}[t][q] \pluseq p$}\label{line:update_log}
    \EndFor
  \EndFor
\EndFor

\State{$ Z = \sum_{q \in Q} \mathsf{forward}[n][q] f[q]$}

\For{$t=0, \ldots, n-1$}
  \For{$q, q' \in Q$}
      \State{$\alpha = \mathsf{forward}[t][q]$} \Comment{Expected counts}
      \State{$\beta = \mathsf{backward}[t+1][q']$}
      \State{$\mathsf{count}[q,q']\pluseq \alpha \times \delta[w][q, q'] \times \beta /  Z$}\label{line:update2}

  \EndFor
\EndFor

\end{algorithmic}
\caption{Forward-Backward algorithm (row-major).}
\label{alg:forward_backward_csr}
\end{algorithm}

If the transition matrices $\delta[a]$ are stored in compressed sparse row (CSR) format, which enables efficient traversal of a matrix in row-major order, then these algorithms can be written out as Algorithm \ref{alg:forward_backward_csr} for the forward-backward algorithm and \ref{alg:forwardcoo} for Viterbi. (Using compressed sparse columns (CSC) format, the loop over $q'$ would be outside the loop over $q$, which is perhaps the more common way to implement these algorithms.)

\section{Parallel Algorithm}
\label{sec:implementation}

Our parallel implementation is based on Algorithm~\ref{alg:forwardcoo} for Viterbi and Algorithm \ref{alg:forward_backward_csr} for forward-backward, but parallelizes the loop over $t$, that is, over the transitions on symbol $w_t$.
The transitions are stored in CSR/COO format as described above for Algorithm~\ref{alg:forwardcoo}. The $S$, $T$, and $P$ arrays are stored on the GPU in global memory; the $R$ and $O$ arrays are kept on the host. For each input symbol $a$, the transitions on $S$ and $T$ are sorted first by source state and then by target state; this improves memory locality slightly. For the forward-backward algorithm, sorting by target improves the performance for the backward pass since the input is read from right to left.

For each input symbol $w_t$, one thread is launched per transition, that is, for each nonzero entry of the transition matrix $\delta[w_t]$. Equivalently, one thread is launched for each transition $k$ such that $R[w_t] \leq k < R[w_t+1]$, for a total of $R[w_{t}+1]-R[w_{t}]$ threads. Each thread looks up $q = S[k], q' = T[k]$ and computes its corresponding operation.

For example, in Figure \ref{fig:coo}, input word ``le" has index 0; since $R[0]=0$ and $R[1]=2$, two threads are launched, one for $k=0$ (that is, $0 \xrightarrow{\text{le}:\text{the}/0.48} 1$) and one for $k=1$ (that is, $0 \xrightarrow{\text{le}:\text{a}/0.08} 2$).


\subsection{Viterbi}
At the time of computing a transition $\delta[w_t][q,q']$, if the probability (at line~\ref{line:max} in Algorithm \ref{alg:forwardcoo})  is higher than $\alpha[t][q']$, we store the probability in $\alpha[t][q']$. Because this update potentially involves concurrent reads and writes at the same memory location, we use an atomic max operation (defined as \verb|atomicMax| on the NVIDIA toolkit). However, \verb|atomicMax| is not defined for floating-point values.
Additionally, this update needs to store a back-pointer ($k$) that will be used afterwards to reconstruct the highest-probability path. The problem is that the \verb|atomicMax| provided by NVIDIA can only update a single value atomically.

We solve both problems with a trick: pack the Viterbi probability and the back-pointer into a single 64-bit integer, with the probability in the higher 32 bits and the back-pointer in the lower 32 bits. In IEEE 754 format, the mapping between nonnegative real numbers and their bit representations (viewed as integers) is order-preserving, so a max operation on this packed representation updates both the probability and the back-pointer simultaneously.

The reconstruction of the Viterbi path is not parallelizable, but is done on the GPU to avoid copying $\alpha$ back to the host avoiding a slowdown. This generates a sequence of transition indices, which is moved back to the host. There, the output symbols can be looked up in array $O$.

\subsection{Forward-Backward}
The forward and backward algorithms \ref{alg:forward_backward_csr}
are similar to the Viterbi algorithm, but do not need to keep back-pointers. In the forward algorithm, when a transition $\delta[w_t][q,q']$ is processed, we update the sum of probabilities reaching state $q'$ in $\mathsf{forward}[t+1][q']$. Likewise, in the backward algorithm, we update the sum of probabilities starting from $q$ in $\mathsf{backward}[t][q]$. Both passes require atomic addition operations, but because we use log-probabilities to avoid underflow, the atomic addition must be implemented as:
\begin{equation}
\log(\exp {a}+\exp {b}) = b+\logonep( \exp (a - b)),
\label{eq:logsumexp}
\end{equation}
assuming $a \leq b$ and where $\logonep(x) = \log (1+x)$, a common function in math libraries which is more numerically stable for small $x$.

We implemented an atomic version of this log-add-exp operation. The two transcendentals are expensive, but CUDA's fast math option (\verb|-use_fast_math|) speeds them up somewhat by sacrificing some accuracy.

\section{Other Approaches}

\subsection{Parallel prefix sum}

We have already mentioned a line of work begun by 
\cite{ladner+fischer:1980} for unweighted, nondeterministic finite automata, and continued by \cite{hillis} and \cite{Mytkowicz} for unweighted, deterministic finite automata. These approaches use \emph{parallel prefix sum} to compute the weight (\ref{eq:weight}), multiplying each adjacent pair of matrices in parallel and repeating until all the matrices have been multiplied together.

This approach could be combined with ours; we leave this for future work. A possible issue is that matrix-vector products are replaced with slower matrix-matrix products. Another is that prefix sum might not be applicable in a more general setting -- for example, if a FST is composed with an input lattice rather than an input string.

\subsection{Matrix libraries}

The formulation of the Viterbi and forward-backward algorithms as a sequence of matrix multiplications suggests two possible easy implementation strategies. First, if transition matrices are stored as dense matrices, then the forward algorithm becomes identical to forward propagation through a rudimentary recurrent neural network. 
Thus, a neural network toolkit could be used to carry out this computation on a GPU. However, in practice, because our transition matrices are sparse, this approach will probably be inefficient.

Second, off-the-shelf libraries exist for sparse matrix/vector operations, like cuSPARSE.\footnote{\url{http://docs.nvidia.com/cuda/cusparse/}} However, such libraries do not allow redefinition of the addition and multiplication operations, making it difficult to implement the Viterbi algorithm or use log-probabilities. Also, parallelization of sparse matrix/vector operations depends heavily on the sparsity pattern \cite{bell+garland:2008}, so that an off-the-shelf library may not provide the best solution for finite-state models of language. We test this approach below and find it to be several times slower than a non-GPU implementation.

\begin{figure}
\begin{center}
\begin{tabular}{cc}
\scalebox{0.75}{%
\begin{tikzpicture}[baseline=0]
\node[initial,state](bos) at (0,0) {\small\bos};
\node[state](the) at (1,1.25) {\small\texttt{the}};
\node[state](a) at (1,-1.25) {\small\texttt{a}};
\node[state](cat) at (2,0) {\small\texttt{cat}};
\node[accepting,state,inner sep=0](eos) at (4,0) {\small\eos};
\draw (bos) edge node[label] {\texttt{the}/0.8} (the);
\draw (bos) edge node[label] {\texttt{a}/0.2} (a);
\draw (the) edge node[label] {\texttt{cat}/1} (cat);
\draw (a) edge node[label] {\texttt{cat}/1} (cat);
\draw (cat) edge node[label] {\texttt{\eos}/1} (eos);
\end{tikzpicture}} &
\scalebox{0.75}{%
\begin{tikzpicture}[baseline=0]
\node[initial,accepting,state] (q) {};
\draw (q) edge[out=120,in=96,loop] node[label]{\texttt{the}:\texttt{le}/0.6} (q);
\draw (q) edge[out=48,in=24,loop] node[label]{\texttt{a}:\texttt{le}/0.4} (q);
\draw (q) edge[out=-48,in=-24,loop] node[label,below]{\texttt{cat}:\texttt{chat}/1} (q);
\draw (q) edge[out=-120,in=-96,loop] node[label,below]{\eos:\eos/1} (q);
\end{tikzpicture}} \\
(a) & (b)
\end{tabular} \\[1ex]
\scalebox{0.75}{%
\begin{tikzpicture}[node distance=1.67cm,baseline=0]
\node[initial,state] (0) {};
\node[state,right of=0] (1) {};
\node[state,right of=1] (2) {};
\node[accepting,state,right of=2] (3) {};
\draw (0) edge node[label] {\texttt{le}} (1);
\draw (1) edge node[label] {\texttt{chat}} (2);
\draw (2) edge node[label] {\eos} (3);
\end{tikzpicture}} \\
(c)
\end{center}
\caption{Example automata/transducers for  (a) language model (b) translation model (c) input sentence. These three composed together form the transducer in Figure~\ref{fig:fst_example}.}
\label{fig:components}
\end{figure}
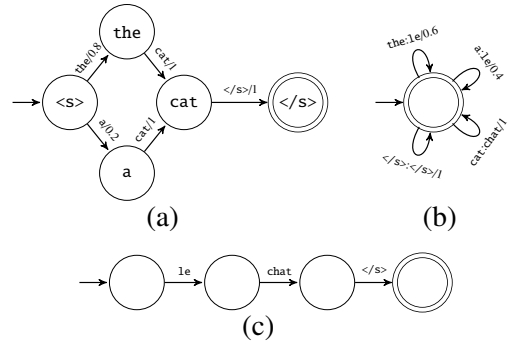

\section{Experiment}

\subsection{Setup}
To test our algorithm, we constructed a FST for rudimentary French-to-English translation. We trained different unsmoothed bigram language models on 1k/10k/100k/150k lines of French-English parallel data from the Europarl corpus and converted it into a finite automaton (see Figure~\ref{fig:components}a for a toy example). 
 
GIZA++ was used to word-align the same data and generate word-translation tables $P(f\mid e)$ from the word alignments, as in lexical weighting \cite{koehn+al:2003}. We converted this table into a single-state FST (Figure~\ref{fig:components}b). The language model automaton and the translation table transducer were intersected to create a transducer similar to the one in Figure \ref{fig:fst_example}.


For more details about the transducers (number of nodes, edges, and percentage of non-zero elements on the transducer) see Table \ref{tab:fst_comp}.

We tested on a subset of 100 sentences from the French corpus with lengths of up to 80 words. For each experimental setting, we ran on this set 1000 times and report the total time. Our experiments were run on three different systems: (1) a system with an Intel Core i7-4790 8-core CPU and an NVIDIA Tesla K40c GPU, (2) a system with an Intel Xeon E5 16-core CPU and an NVIDIA Titan X GPU, and (3) a system with an Intel Xeon E5 24-core CPU and an NVIDIA Tesla P100 GPU. 

\subsection{Baselines}

We compared against the following baselines:

\begin{trivlist}
\item\textbf{Carmel} is an FST toolkit developed at USC/ISI.\footnote{\url{https://github.com/graehl/carmel}}

\item\textbf{OpenFST} is a FST toolkit developed by Google as an open-source successor of the AT\&T Finite State Machine library \cite{openfst}. For compatibility, our implementations read the OpenFST/AT\&T text file format.

\item\textbf{Our serial implementation}  Algorithm~\ref{alg:forwardcoo} for Viterbi and Algorithm \ref{alg:forward_backward_csr} for forward-backward.

\item\textbf{cuSPARSE} was used to implement the forward algorithm, using CSR format instead of COO for transition matrices. Since we can't redefine addition and multiplication, we could not implement the Viterbi algorithm. To avoid underflow, we rescaled the vector of forward values at each time step and kept track of the log of the scale in a separate variable.
\end{trivlist}
To be fair, it should be noted that Carmel and OpenFST are much more general than the other implementations listed here. Both perform FST composition in order to decode an input string adding another layer of complexity to the process. The timings for OpenFST and Carmel on Table \ref{tab:diff-results} include composition

\subsection{Results}

\begin{table*}[t]
\centering
\begin{tabular}{@{}ll|*{8}{c|}@{}}
&  & \multicolumn{8}{|c|}{Training size (lines)} \\
&  & \multicolumn{2}{|c|}{1000} & \multicolumn{2}{|c|}{10000} & \multicolumn{2}{|c|}{100000} &
\multicolumn{2}{|c|}{150000} \\
 \hline
 Method & Hardware & Time & Ratio & Time & Ratio & Time & Ratio & Time & Ratio\\
  \hline
   OpenFST & Core i7 & 42.06 & 1.9 & 2547 & 87 & 313800 & 4085 & 626700 & 6093\\
   Carmel & Core i7 & 195.9 & 9.1 & 7652 &  263 & 224500  & 2923 & 376400 & 3659\\
   our serial & Core i7 & 10.99 & 0.5 &  44.16 & 1.5 & 374.2 & 4.9 & 534.9 & 5.2\\
   our serial & Xeon E5 & \textbf{10.84} & \textbf{0.5} & 42.05& 1.4 & 375.3& 4.9 & 529.6 & 5.1\\
   \hline
   our MPI & 4-core Core i7 & 194.0 & 9.0 & 581.2 & 20 & 1849 & 24 & 2243 & 21\\
   our parallel & K40 & 27.52 & 1.3 & 38.26 & 1.3 & 116.5 &1.5 & 131.1 & 1.3\\
   our parallel & Titan X  &25.05& 1.2 & 33.92 & 1.2 & 94.07 & 1.2 & 121.6 & 1.2 \\
   our parallel & Tesla P100 & 21.49 & 1.0 & \textbf{29.04} & \textbf{1.0} & \textbf{76.79} & \textbf{1.0} & \textbf{102.9}& \textbf{1.0}\\
  
\end{tabular}

\caption{\label{tab:diff-results}Our GPU implementation of the Viterbi algorithm outperforms all others tested on the medium and large FSTs. Times (in seconds) are for decoding a set of 100 examples 1000 times using Viterbi. Ratios are relative to our parallel algorithm on the Tesla P100. 
}
\end{table*}

\begin{table}
\begin{center}
\begin{tabular}{@{}l|rrrr@{}}
& \multicolumn{4}{c}{Training size (lines)} \\
method & 1k & 10k & 100k & 150k \\
\hline
 cuSPARSE forward & 646 & 1846 & 3555 & 5948\\
 serial forward & 36 & 251 & 2297 & 3346\\
 parallel forward & \textbf{17} & \textbf{37} & \textbf{236} & \textbf{327}\\
 \hline
 serial backward & \textbf{13} & 248 & 3585 & 5303\\
 parallel backward & 43 & \textbf{80} & \textbf{644} & \textbf{1070}\\
 \hline
 serial combined & \textbf{47} & 534 & 6065 &8790 \\
 parallel combined  & 60 & \textbf{120} & \textbf{1111} & \textbf{1773}\\
\hline

\end{tabular}
\end{center}
\caption{\label{tab:for_back_comp}Our GPU implementations of the forward and backward algorithms, and forward+backward+expected counts combined, outperform all others tested, on the medium and large FSTs. Times (in seconds) are for processing 100 examples 1000 times, on a Core i7 and K40.}
\end{table}

\begin{table}
\begin{center}
\begin{tabular}{l|rrr}
Training size & States & Transitions & Non-zero \\
\hline
 1000 & 3505 & 443527  & 3.6\%\\
 \hline
  10000 & 11644 & 6792487 & 5.0\% \\
\hline
  100000 & 33125 & 95381368 &  8.7\%\\
\hline
  150000 & 39420 & 150971615 &  9.7\%\\
\hline
\end{tabular}
\end{center}
\caption{FST Comparison. This table shows the number of states, edges, and percent of non zero elements of the transducers created using 1k/10k/100k/150k examples. }
\label{tab:fst_comp}
\end{table}

\begin{figure}
\begin{center}
\includegraphics[width=\hsize]{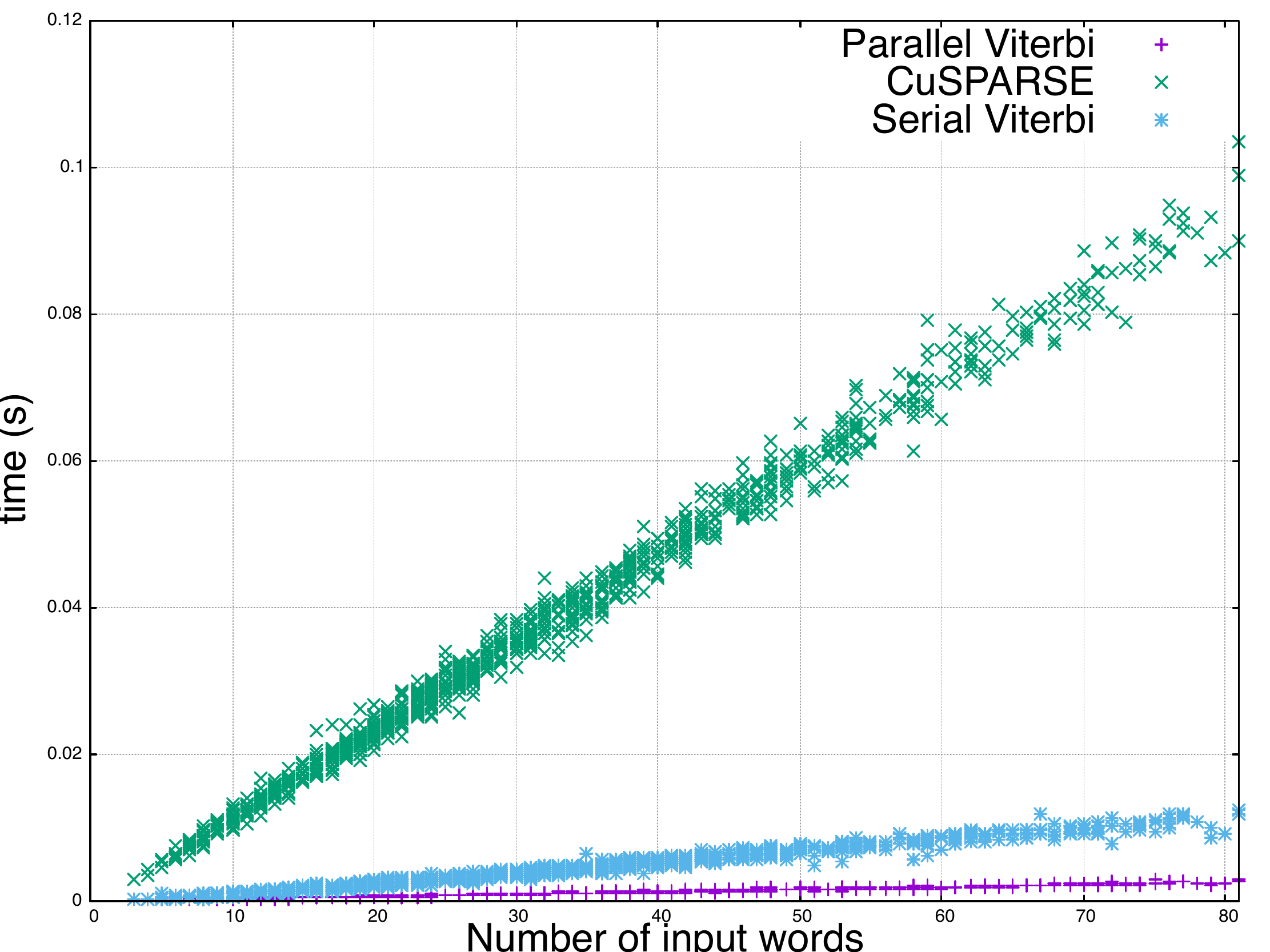}
\end{center}
\caption{Viterbi decoding times for 1000 individual test sentences compared for our serial, parallel, and cuSPARSE implementations (Titan X).}
\label{fig:senttimes}
\end{figure}

Table~\ref{tab:diff-results} shows the overall performance of our Viterbi algorithm and the baseline algorithms. Our parallel implementation does worse than our serial implementation when the transducer used is small (presumably due to the overhead of kernel launches and memory copies), but the speedups increase as the size of the transducer grows, reaching a speedup of 5x. The forward-backward algorithm with expected counts obtains a 5x speedup over the serial code on the largest transducer (See Table \ref{tab:for_back_comp}).


CuSPARSE does significantly worse than even our serial implementation; presumably, it would have done better if the transition matrices of our transducers were sparser.

Figure~\ref{fig:senttimes} shows decoding times for three algorithms (our serial and parallel Viterbi, and cuSPARSE forward) on individual sentences. It can be seen that all three algorithms are roughly linear in the sentence length.

Viterbi is faster than either the forward or backward algorithm across the board. This is because the latter need to add log-probabilities (lines \ref{line:update1} and \ref{line:update_log} of Algorithm \ref{alg:forward_backward_csr}), which involves expensive calls to transcendental functions.

\subsection{Comparison across GPU architectures}

Table \ref{tab:diff-results} compares the performance of the Kepler-based K40, where we did most of our experiments, with the Maxwell-based Titan X and the  Pascal-based Tesla P100. The performance improvement is due to different factors, such as a larger number of active thread blocks per streaming multiprocessor on a GPU architecture, the grid and block size selected to run the kernels, and memory management on the GPU. After the release of the Kepler architecture, the Maxwell architecture introduced an improved workload balancing, reduced arithmetic latency, and faster atomic operations. The Pascal architecture allows speedups over all the other architectures by introducing an increased floating point performance, faster data movement performance (NVLink), larger and more efficient shared memory, and improved atomic operations. Also, SMs on the pascal architecture are more efficient allowing speedups larger speedups than its predecessors. Our parallel implementations were compiled using architecture specific flags (\verb|-arch=compute_XX|) to take full advantage of the architectural enhancements described in this section.

\subsection{Comparison against a multi-core implementation}
Table \ref{tab:diff-results} shows how our parallel implementation on a GPU compares against a multi-core version of our serial Viterbi algorithm implemented in MPI. We chose MPI since it supports distributed and shared memory unlike OpenMP that supports shared memory only. Results show that a multi-core implementation of the algorithm leads to slower performance than the serial code due to the communication and synchronization overhead. Several cores must transfer information frequently and synchronize all messages on a single core. GPUs perform better than multi-core in this case since all the memory is already on the graphics card and the cost of using global memory on the GPU is lower than synchronizing and sharing data between cores.

\section{Conclusion}

We have shown that our algorithm outperforms several serial implementations (our own serial implementation on a Intel Core i7 and Xeon E machines, Carmel and OpenFST) as well as a GPU implementation using cuSPARSE. 

A system with newer and faster cores might achieve higher speedups than a GPU on smaller datasets. However,building a multi-core system that beats a GPU setup can be more expensive. For example, a 16 core Intel Xeon E5-2698 V3 can cost 3,500 USD \cite{bogoychevn}. Newer GPU models offer previous generation CPU's the opportunity to obtain speedups for a lower price (Titan X GPUs sell cheaper than Xeon E5 setups at US\$1,200). Speeding up computation on a GPU would allow users to speed up applications cheaper without investing on a newer multi-core system.

Our implementation has been open-sourced and is available online. \footnote{\url{https://bitbucket.org/aargueta2/parallel-decoding}} In the future, we plan to expand this software into a toolkit that includes other algorithms needed to run a full machine translation system.

\section*{Acknowledgements}
This research was supported in part by a gift of a Tesla K40c GPU card from NVIDIA Corporation.

\bibliography{eacl2017}

\begin{thebibliography}{}

\bibitem[\protect\citename{Allauzen \bgroup et al.\egroup }2007]{openfst}
Cyril Allauzen, Michael Riley, Johan Schalkwyk, Wojciech Skut, and Mehryar
  Mohri.
\newblock 2007.
\newblock {OpenFst}: A general and efficient weighted finite-state transducer
  library.
\newblock In {\em Proc.~International Conference on Implementation and
  Application of Automata (CIAA 2007)}, pages 11--23.

\bibitem[\protect\citename{Bell and Garland}2008]{bell+garland:2008}
Nathan Bell and Michael Garland.
\newblock 2008.
\newblock Efficient sparse matrix-vector multiplication on {CUDA}.
\newblock Technical Report NVIDIA Technical Report NVR-2008-004, NVIDIA
  Corporation.

\bibitem[\protect\citename{Bogoychev and Lopez}2016]{bogoychevn}
Nikolay Bogoychev and Adam Lopez.
\newblock 2016.
\newblock N-gram language models for massively parallel devices.
\newblock In {\em Proc. ACL}, pages 1944--1953.

\bibitem[\protect\citename{Canny \bgroup et al.\egroup }2013]{canny2013}
John Canny, David Hall, and Dan Klein.
\newblock 2013.
\newblock A multi-teraflop constituency parser using {GPUs}.
\newblock In {\em Proc. EMNLP}, pages 1898--1907.

\bibitem[\protect\citename{Chong \bgroup et al.\egroup }2009]{chong2009f}
Jike Chong, Ekaterina Gonina, Youngmin Yi, and Kurt Keutzer.
\newblock 2009.
\newblock A fully data parallel {WFST}-based large vocabulary continuous speech
  recognition on a graphics processing unit.
\newblock In {\em Proc. INTERSPEECH}, pages 1183--1186.

\bibitem[\protect\citename{Eisner}2002]{eisner:2002}
Jason Eisner.
\newblock 2002.
\newblock Parameter estimation for probabilistic finite-state transducers.
\newblock In {\em Proc. ACL}, pages 1--8.

\bibitem[\protect\citename{Hall \bgroup et al.\egroup }2014]{hall}
David Hall, Taylor Berg-Kirkpatrick, and Dan Klein.
\newblock 2014.
\newblock Sparser, better, faster {GPU} parsing.
\newblock In {\em Proc. ACL}, pages 208--217.

\bibitem[\protect\citename{He \bgroup et al.\egroup }2013]{he2013}
Hua He, Jimmy Lin, and Adam Lopez.
\newblock 2013.
\newblock Massively parallel suffix array queries and on-demand phrase
  extraction for statistical machine translation using gpus.
\newblock In {\em Proc. NAACL HLT}, pages 325--334.

\bibitem[\protect\citename{Hillis and Steele}1986]{hillis}
W.~Daniel Hillis and Guy~L. Steele, Jr.
\newblock 1986.
\newblock Data parallel algorithms.
\newblock {\em Communications of the ACM}, 29(12):1170--1183.

\bibitem[\protect\citename{Kim \bgroup et al.\egroup }2012]{kim2012}
Jungsuk Kim, Jike Chong, and Ian~R Lane.
\newblock 2012.
\newblock Efficient on-the-fly hypothesis rescoring in a hybrid gpu/cpu-based
  large vocabulary continuous speech recognition engine.
\newblock In {\em Proc. INTERSPEECH}, pages 1035--1038.

\bibitem[\protect\citename{Koehn \bgroup et al.\egroup }2003]{koehn+al:2003}
Philipp Koehn, Franz~Josef Och, and Daniel Marcu.
\newblock 2003.
\newblock Statistical phrase-based translation.
\newblock In {\em Proc. NAACL HLT}, pages 48--54.

\bibitem[\protect\citename{Kumar \bgroup et al.\egroup }2005]{kumar+al:2005}
Shankar Kumar, Yonggang Deng, and William Byrne.
\newblock 2005.
\newblock A weighted finite state transducer translation template model for
  statistical machine translation.
\newblock {\em J.~Natural Language Engineering}, 12(1):35--75.

\bibitem[\protect\citename{Ladner and Fischer}1980]{ladner+fischer:1980}
Richard~E. Ladner and Michael~J. Fischer.
\newblock 1980.
\newblock Parallel prefix computation.
\newblock {\em J. ACM}, 27(4):831--838.

\bibitem[\protect\citename{Lafferty \bgroup et al.\egroup
  }2001]{lafferty+al:2001}
John~D. Lafferty, Andrew McCallum, and Fernando C.~N. Pereira.
\newblock 2001.
\newblock Conditional random fields: Probabilistic models for segmenting and
  labeling sequence data.
\newblock In {\em Proc. ICML}, pages 282--289.

\bibitem[\protect\citename{Li \bgroup et al.\egroup }2014]{li2014}
Rongchun Li, Yong Dou, and Dan Zou.
\newblock 2014.
\newblock Efficient parallel implementation of three-point viterbi decoding
  algorithm on {CPU}, {GPU}, and {FPGA}.
\newblock {\em Concurrency and Computation: Practice and Experience},
  26(3):821--840.

\bibitem[\protect\citename{Merrill \bgroup et al.\egroup }2012]{Merril}
Duane Merrill, Michael Garland, and Andrew Grimshaw.
\newblock 2012.
\newblock Scalable {GPU} graph traversal.
\newblock In {\em Proc.~17th ACM SIGPLAN Symposium on Principles and Practice
  of Parallel Programming (PPoPP)}, pages 117--128.

\bibitem[\protect\citename{Mohri \bgroup et al.\egroup }2002]{mohri+al:2002}
Mehryar Mohri, Fernando C.~N. Pereira, and Michael Riley.
\newblock 2002.
\newblock Weighted finite-state transducers in speech recognition.
\newblock {\em Computer Speech and Language}, 16(1):69--88.

\bibitem[\protect\citename{Mohri}2009]{mohri:2009}
Mehryar Mohri.
\newblock 2009.
\newblock Weighted automata algorithms.
\newblock In Manfred Droste, Werner Kuich, and Heiko Vogler, editors, {\em
  Handbook of Weighted Automata}, pages 213--254. Springer.

\bibitem[\protect\citename{Mytkowicz \bgroup et al.\egroup }2014]{Mytkowicz}
Todd Mytkowicz, Madanlal Musuvathi, and Wolfram Schulte.
\newblock 2014.
\newblock Data-parallel finite-state machines.
\newblock In {\em Proc.~Architectural Support for Programming Languages and
  Operating Systems (ASPLOS)}, March.

\bibitem[\protect\citename{Narasiman \bgroup et al.\egroup
  }2011]{narasiman2011}
Veynu Narasiman, Michael Shebanow, Chang~Joo Lee, Rustam Miftakhutdinov, Onur
  Mutlu, and Yale~N Patt.
\newblock 2011.
\newblock Improving {GPU} performance via large warps and two-level warp
  scheduling.
\newblock In {\em Proc. IEEE/ACM International Symposium on Microarchitecture},
  pages 308--317.

\bibitem[\protect\citename{Peng \bgroup et al.\egroup }2016]{peng2016}
Hao Peng, Rongke Liu, Yi~Hou, and Ling Zhao.
\newblock 2016.
\newblock A {Gb/s} parallel block-based viterbi decoder for convolutional codes
  on gpu.
\newblock {\em arXiv preprint arXiv:1608.00066}.

\bibitem[\protect\citename{{Theano Development Team}}2016]{theano}
{Theano Development Team}.
\newblock 2016.
\newblock {Theano}: A {Python} framework for fast computation of mathematical
  expressions.
\newblock {\em arXiv}, abs/1605.02688, May.

\end{thebibliography}
\bibliographystyle{eacl2017}

\end{document}